\pdfoutput=1
\documentclass[conference]{IEEEtran}
\IEEEoverridecommandlockouts

\makeatletter
\g@addto@macro\normalsize{%
  \setlength\abovedisplayskip{4pt plus 2pt minus 2pt}%
  \setlength\belowdisplayskip{4pt plus 2pt minus 2pt}%
  \setlength\abovedisplayshortskip{2pt plus 2pt minus 1pt}%
  \setlength\belowdisplayshortskip{2pt plus 2pt minus 1pt}%
}
\makeatother
\usepackage{amsmath,amssymb,amsfonts}
\usepackage{bm}
\usepackage{mathtools}
\usepackage{graphicx}
\usepackage[caption=false,font=footnotesize]{subfig} 
\usepackage{booktabs}
\usepackage{multirow}
\usepackage{array}
\usepackage{makecell}
\usepackage{tabularx}
\usepackage{siunitx}
\usepackage{algorithm}
\usepackage{algpseudocode}
\usepackage{cite}
\usepackage{url}
\usepackage[hidelinks]{hyperref}
\usepackage{xcolor}
\usepackage{microtype}
\def\BibTeX{{\rm B\kern-.05em{\sc i\kern-.025em b}\kern-.08em
    T\kern-.1667em\lower.7ex\hbox{E}\kern-.125emX}}
\begin{document}

\title{Dose-Aware Cold Diffusion with Physics Consistency for Generalizable Low-Dose CT Reconstruction}
\author{
\IEEEauthorblockN{
Md Imam Ahasan\textsuperscript{1}
Guangchao Yang\textsuperscript{1*}
A. F. M. Abdun Noor\textsuperscript{2}
S. M. Hasan Mahmud\textsuperscript{2}
Md Mahfuzur Rahman\textsuperscript{1}
}

\IEEEauthorblockA{
\textsuperscript{1}College of Computer Science, Chongqing University, Chongqing, China \\
\textsuperscript{2}Department of Software Engineering, Daffodil International University, Dhaka, Bangladesh \\
}
\thanks{*Corresponding author: Guangchao Yang (gchao\_yang@cqu.edu.cn).}
\thanks{This work was supported by the National Natural Science Foundation of China under Grant 62572086.}
}
\maketitle
\begin{abstract}
Reducing radiation dose in computed tomography significantly degrades image quality and poses challenges for accurate and clinically reliable reconstruction. While recent approaches have shown promise for low-dose CT, they often struggle to generalize across continuous and previously unseen dose levels, leading to artifacts and loss of anatomical detail. To address these limitations, we propose Dose-Aware Cold Diffusion (DACD), a physics-consistent reconstruction framework that explicitly models radiation dose as a continuous latent factor within a cold diffusion process. The proposed DACD framework integrates image-based dose-aware perception, multi-scale structural prior extraction, and dose-calibrated step allocation to adaptively guide the denoising trajectory. In addition, an iterative forward-backprojection correction is incorporated into the reverse refinement process to enforce projection-domain data consistency. Extensive experiments on three public benchmarks, including Mayo-2020, Mayo-2016, and LoDoPaB-CT, demonstrate that DACD consistently outperforms state-of-the-art diffusion-based and physics-guided methods in both quantitative accuracy and visual fidelity, particularly under ultra-low-dose conditions. The results show that DACD achieves robust generalization across a continuous range of dose levels, including those unseen during training.
\end{abstract}


\section{Introduction}
\label{sec:introduction}
\begin{figure}[!ht]
\centering
\includegraphics[width=1\linewidth]{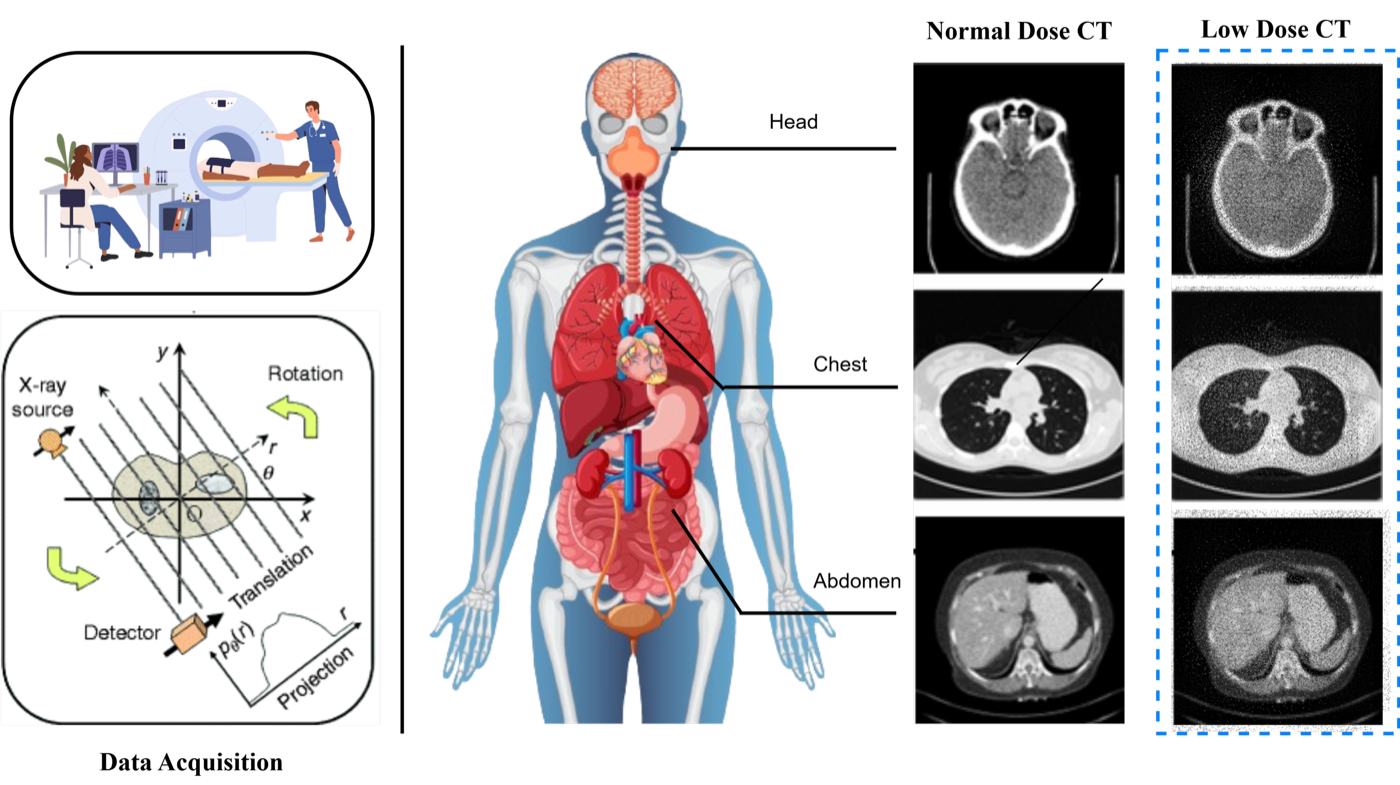}
\caption{Overview of CT imaging workflow and dose effects. Varying X-ray tube currents alter photon statistics and noise levels in reconstructed images, highlighting the importance of dose-aware reconstruction.}
\label{fig:concept}
\end{figure}

Computed tomography (CT) is central to clinical diagnosis, screening, and image-guided intervention. However, X-ray radiation exposure raises persistent safety concerns~\cite{smith2009radiation,sodickson2009recurrent}. In practice, dose reduction is commonly achieved by lowering tube current, which reduces photon statistics and yields increased noise and pronounced streak artifacts in reconstructed images. Fig.~\ref{fig:concept} illustrates how decreasing tube current induces distinct degradation patterns in low-dose CT (LDCT) reconstructions. Prior LDCT methods use model-based iterative and classical post-processing techniques~\cite{wang2006penalized,ma2011low,xie2017robust}, operating in either the sinogram domain to denoise projections before FBP or directly in the image domain. Representative techniques include penalized weighted least squares~\cite{zha2023rank}, bilateral filtering~\cite{khosla2020supervised}, and structural filtering~\cite{peebles2023scalable}, which aim to preserve structural fidelity under noisy acquisition or limited sampling.
Deep learning (DL) has also been widely adopted for LDCT denoising~\cite{chen2017low, chi2024low}. Most existing DL approaches are trained for a single dose level or a specific anatomical region, limiting their ability to handle heterogeneous noise characteristics and anatomical variability across protocols and scanners~\cite{chen2023ascon}. This is problematic in clinical deployment, where dose settings and reconstruction pipelines vary with patient factors, target anatomy, and manufacturer-specific configurations, even under the same nominal parameters~\cite{gao2025noise}. Training separate models per dose or anatomy is feasible but computationally expensive and difficult to scale.
\begin{figure*}[!ht]
    \centering
    \includegraphics[width=1\linewidth]{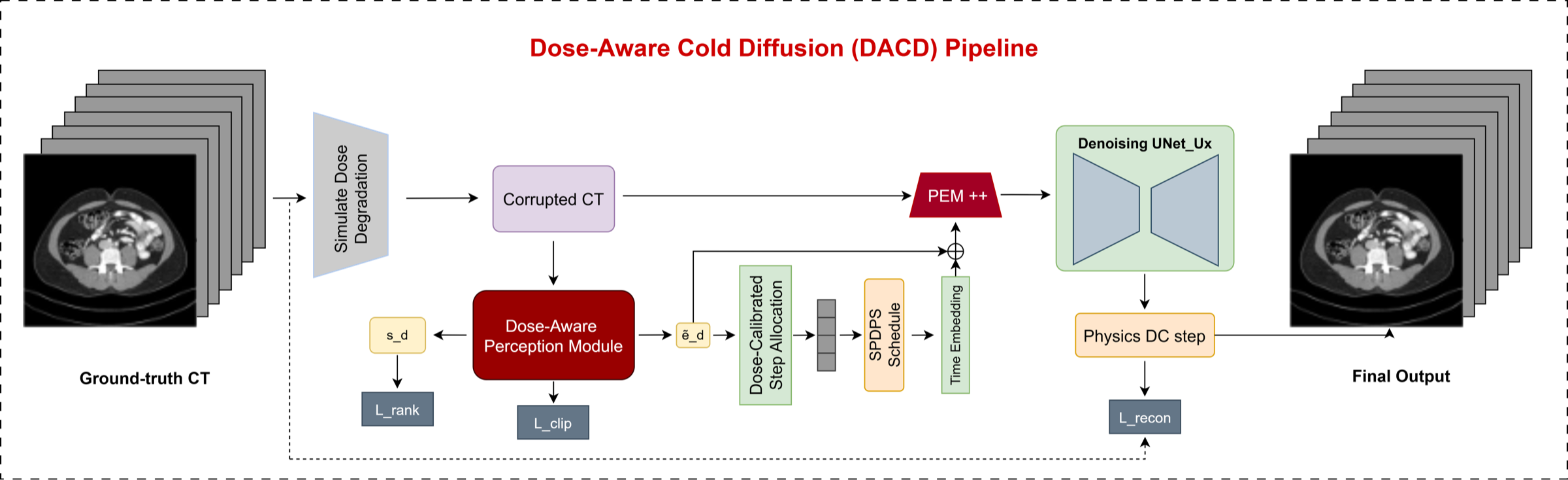}
    \caption{Overview of the Dose-Aware Cold Diffusion (DACD) framework. The model combines physics-based consistency with adaptive diffusion denoising, guided by dose-aware embeddings that modulate reconstruction according to noise severity.}
    \label{fig:dacd_overview}
\end{figure*}
Denoising diffusion probabilistic models (DDPMs) have recently achieved strong performance in image generation~\cite{ho2020denoising, nichol2021improved, choi2021ilvr}, with stable training and high synthesis quality~\cite{nichol2021improved, dhariwal2021diffusion, M_ller_Franzes_2023, saharia2022palette}. DDPMs learn to iteratively remove Gaussian noise through a denoising network, but standard sampling is slow. To accelerate diffusion-based CT reconstruction, fast ordinary differential equation (ODE) solvers have been used to reduce sampling steps~\cite{xia2022low,lu2022dpm}. Building on cold diffusion, Gao \textit{et al.}~(2023) proposed the contextual error-modulated generalized diffusion (CoreDiff) model for LDCT denoising~\cite{bansal2023cold, gao2023corediff, mardani2023variational}. Subsequent work incorporated manifold constraints and prior features to improve score-based reconstruction under ultra-sparse sampling~\cite{chung2022improving, chung2023decomposed, chung2023solving}, and Dou \textit{et al.}~(2024) introduced a sequential Monte Carlo strategy for asymptotically accurate Bayesian posterior sampling with strong zero-shot performance~\cite{dou2024diffusion}. Liu \textit{et al.}~(2024) learn a general CT denoiser across multiple degradations~\cite{liu2024residual}, but it lacks adaptive dose perception and requires costly fine-tuning for unseen data; dose-conditional methods improve generalization~\cite{xia2021ct} yet treat dose as discrete and ignore anatomical variability~\cite{mussmann2021organ}, motivating a continuous, anatomy-aware dose-conditioned approach.

In this work, we propose Dose-Aware Cold Diffusion (DACD), a reconstruction framework for LDCT that combines dose-conditioned diffusion modeling with physics-consistent corrections. DACD explicitly encodes dose effects, incorporates handcrafted structural priors, and enforces fidelity to measured sinograms through iterative forward-backprojection corrections within the reverse refinement process, enabling robust generalization to unseen dose levels. Extensive experiments on three benchmark CT datasets demonstrate the effectiveness of the proposed approach. For reproducibility, the implementation of DACD will be made publicly available at \url{https://github.com/imamahasane/DACD}. The primary contributions are:
\begin{itemize}
\item We propose DACD, a unified framework that models radiation dose as a continuous latent factor for generalizable LDCT reconstruction across a wide range of dose conditions.
\item We introduce an in-loop physics consistency mechanism that integrates iterative forward-backprojection correction into each reverse refinement step, improving physical reliability and reducing hallucinated structures.
\item We develop a dose-calibrated step allocation strategy that adapts the number of reverse refinement steps to degradation severity, improving inference efficiency while maintaining quality under ultra-low-dose settings.
\end{itemize}

\section{Related Work}
Diffusion models have recently achieved strong performance in sparse-view and low-dose CT reconstruction by learning powerful iterative denoisers guided by the CT forward model. Cold diffusion methods replace synthetic Gaussian corruption with physically meaningful degradations, enabling direct inversion of CT-specific operators. Several works tailor diffusion to particular degradation types. CvG-Diff models angular subsampling artifacts for sparse-view CT using a degradation operator and efficient multi-phase sampling~\cite{chen2025cross}, while NEED introduces a Poisson-inspired diffusion process and guided refinement to better match photon noise statistics in low-dose CT~\cite{gao2025noise}.

Other approaches pursue broader generalization across acquisition settings. FoundDiff leverages dose- and anatomy-aware pretraining to adapt a single denoiser to multiple anatomical regions and dose categories~\cite{chen2025founddiff}, and CT-SDM supports reconstruction across different sparse-view sampling rates within one model using projection-domain degradation and stochastic sampling~\cite{yang2025ct}. PrideDiff incorporates sinogram-domain priors in a physics-regularized cold diffusion framework to improve stability and reconstruction quality~\cite{lu2024pridediff}. In contrast to prior work that relies on discrete conditioning or outer-loop data consistency, we model radiation dose as a continuous latent variable that parameterizes the diffusion trajectory and embed forward–backprojection correction inside every reverse step. This tight coupling of dose-aware conditioning and in-loop physics consistency enables robust reconstruction and smooth generalization to unseen dose levels.
\section{Method}
\subsection{Problem Formulation}
Let $x_{gt} \in \mathbb{R}^{H \times W}$ denote the ground-truth attenuation map and $A$ the CT forward projection operator. The ideal noiseless sinogram is
\begin{equation}
    y_{\text{clean}} = A(x_{gt}).
\end{equation}
At a reduced radiation dose level $d \in (0,1]$, photon measurements follow Poisson statistics,
\begin{equation}
    y_{d} \sim \mathrm{Poisson}(d \cdot y_{\text{clean}}),
\end{equation}
where smaller $d$ yields higher relative noise. A standard analytic reconstruction from noisy projections is obtained by filtered backprojection (FBP),
\begin{equation}
    x_{\mathrm{FBP}} = A^{T}_{\mathrm{FBP}}(y_d),
\end{equation}
which is fast but highly sensitive to noise and undersampling. LDCT reconstruction seeks an estimate $\hat{x}$ that is both faithful to the measurements and structurally plausible, typically formulated as
\begin{equation}
    \hat{x} = \arg\min_{x} \; \mathcal{L}\big(A(x), y_d\big) + \lambda \,\mathcal{R}(x),
\end{equation}
where $\mathcal{L}$ enforces data fidelity to the measured sinogram and $\mathcal{R}$ encodes prior knowledge on image structure. The main challenges are to (i) generalize across continuous dose levels, (ii) preserve fine anatomical details under severe noise, and (iii) maintain consistency with the CT forward model. These requirements motivate a dose-aware, physics-consistent reconstruction framework.
\begin{figure}[!ht]
\centering
\includegraphics[width=1\linewidth]{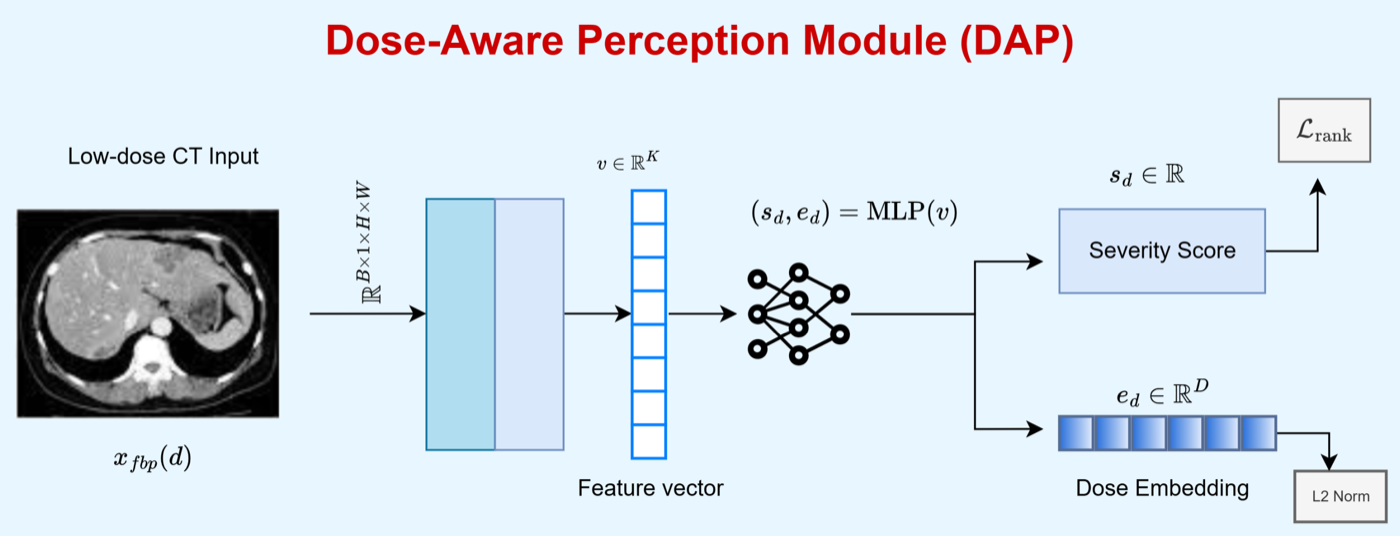}
\caption{Overview of the DAP module. DAP extracts feature embeddings from low-dose CTs to estimate dose severity and generate dose-conditioned representations for adaptive reconstruction.}
\label{fig:DAP}
\end{figure}

\subsection{Dose-Aware Cold Diffusion Framework}
DACD formulates low-dose CT reconstruction as a dose-conditioned, physics-guided iterative refinement process with dose level $d \in (0,1]$. As illustrated in Fig.~\ref{fig:dacd_overview}, the framework combines dose-dependent degradation modeling, dose-aware feature conditioning, structural priors, and projection-domain data consistency within a unified cold diffusion scheme. Instead of adding synthetic Gaussian noise, the forward degradation is generated by the physical low-dose acquisition process.
\paragraph*{Forward Process (Training)}
For a ground-truth attenuation map $x_{gt}$, the clean sinogram is
\begin{equation}
    y_{\text{clean}} = A(x_{gt}).
\end{equation}
A low-dose measurement is simulated by Poisson thinning,
\begin{equation}
    y_d = \text{PoissonThin}(y_{\text{clean}}, d),
\end{equation}
and reconstructed using filtered backprojection (FBP),
\begin{equation}
    x_t(d) = A_{\text{FBP}}^{T}(y_d).
\end{equation}
Varying $d$ produces a sequence of deterministic degraded reconstructions $\{x_t(d)\}$, where smaller $d$ yields more severe degradation. A monotonic mapping between dose level $d$ and diffusion index $t$ defines a physically grounded cold diffusion trajectory that the reverse process inverts.

\textbf{Reverse Process (Inference):} Given a measured low-dose sinogram $y_d$ and its FBP reconstruction, DACD iteratively alternates learned denoising and physics-based correction:
\begin{equation}
    x_{t-1} = U_{\theta}(x_t, c_t)
    - \eta_t\, A^{T}\!\big(A(\tilde{x}_{t-1}) - y_d\big),
\end{equation}
where $\tilde{x}_{t-1}=U_{\theta}(x_t,c_t)$ is the denoised estimate, $U_{\theta}$ is a U-Net denoiser, $\eta_t$ controls the correction strength, and $c_t$ concatenates diffusion-time and dose embeddings. The second term enforces projection-domain data fidelity at every step, coupling the learned prior with the CT forward model.

\textbf{Training Objective:} Training combines image fidelity, dose-order consistency, and representation alignment:
\begin{equation}
    \mathcal{L}_{\text{total}}=
    \mathcal{L}_{\text{img}}
    +\lambda_1\mathcal{L}_{\text{rank}}
    +\lambda_2\mathcal{L}_{\text{contrast}},
\end{equation}
with
\begin{equation}
    \mathcal{L}_{\text{img}}=
    \|x_{t-1}-x_{gt}\|_1
    +\beta\,\|A(x_{t-1})-y_{\text{clean}}\|_2^2.
\end{equation}
This term enforces consistency in both image and projection domains. The ranking and contrastive losses from the dose-aware perception module impose ordinal consistency across dose levels and encourage discriminative dose-aware representations. Together, these constraints promote physically consistent and accurate reconstruction over a continuous range of radiation doses.

\subsection{Dose-to-Step Mapping in Cold Diffusion}
DACD adopts a cold diffusion formulation in which each diffusion step corresponds to a deterministic degradation induced by a specific radiation dose, directly reflecting the physics of low-dose CT. Let $d\in(0,1]$ be the normalized dose fraction ($d=1$ is full dose). We map dose to a degradation index $t\in\{1,\dots,T_{\max}\}$ by
\begin{equation}
    t(d)=\left\lfloor T_{\max}(1-d)\right\rfloor+1,
\end{equation}
so lower doses yield larger indices and more severe degradation. Given full-dose projections $y_{\text{full}}$, low-dose measurements and degraded images are generated as
\begin{equation}
    y_d \sim \mathrm{Poisson}(d\cdot y_{\text{full}}), \qquad
    x_t=\mathcal{A}^{\dagger}(y_d),
\end{equation}
producing a deterministic degradation state for each $t$. The image $x_t$ initializes the reverse refinement. The degradation index $t(d)$ encodes noise severity and is distinct from the number of reverse refinement iterations. The latter is adaptively set by the dose-calibrated step allocation strategy $T(d)$, which assigns more iterations to lower-dose inputs. Any monotonic mapping between $d$ and $t$ preserves dose ordering and yields stable reconstruction due to dose-aware conditioning and in-loop physics-consistent correction, making this mapping a simple and physically interpretable design choice.
\begin{figure}[!ht]
\centering
\includegraphics[width=1\linewidth]{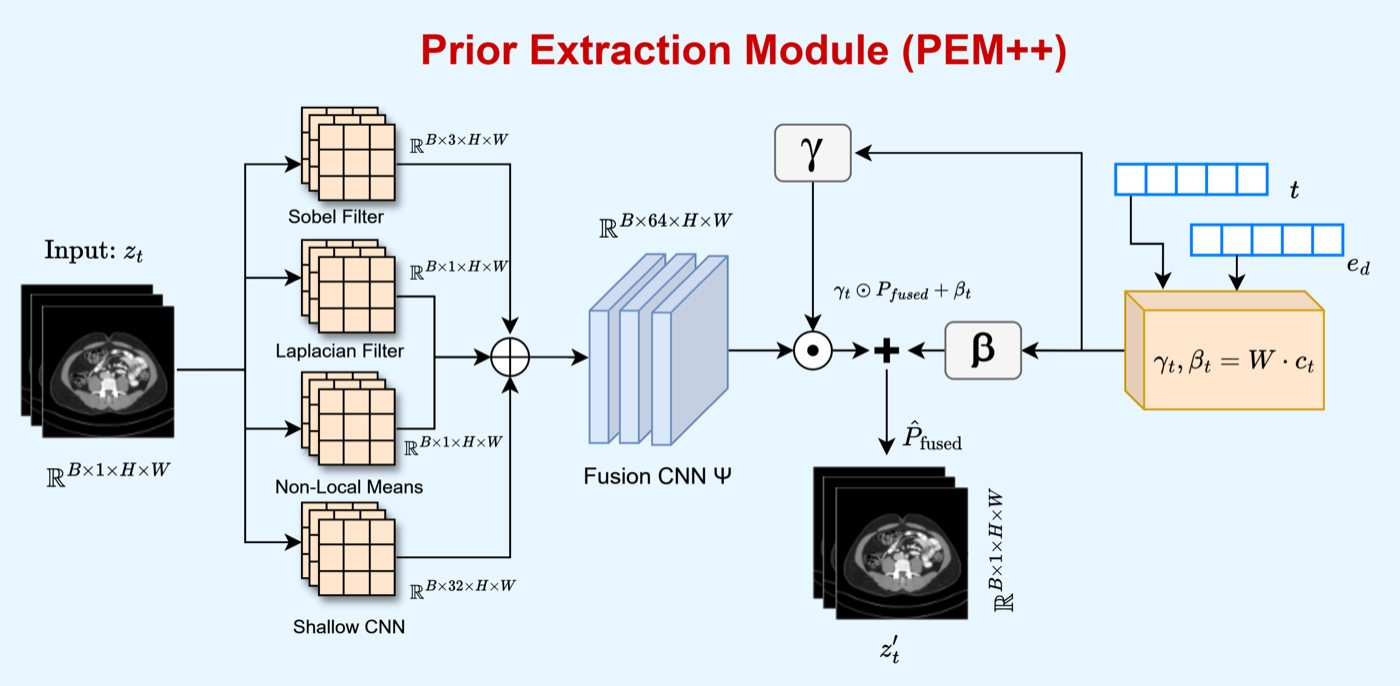}
\caption{Architecture of the Prior Extraction Module (PEM++). The module fuses multi-scale spatial priors from edge, texture, and CNN features, modulated by dose and time embeddings for adaptive reconstruction.}
\label{fig:PEM}
\end{figure}

\subsection{Core Modules}
DACD consists of three components for dose-adaptive reconstruction: the Dose-Aware Perception module (DAP), the Prior Extraction module (PEM++), and the Dose-Calibrated Step Allocation mechanism (DCSA), which respectively enable dose perception, structural prior modeling, and adaptive control of the reverse process.

\textbf{1) Dose-Aware Perception (DAP):} DAP estimates degradation severity and produces a continuous dose-conditioned embedding from the degraded image (Fig.~\ref{fig:DAP}). Given $x_t$, a lightweight encoder extracts latent features
\begin{equation}
    z_t = f_{\theta}(x_t),
\end{equation}
which are mapped by an MLP to a severity score and dose embedding,
\begin{equation}
    (s_d, e_d) = \mathrm{MLP}(z_t),
\end{equation}
with $\ell_2$-normalized $e_d$ for stable conditioning. Ordinal consistency across dose levels is enforced by a ranking loss
\begin{equation}
    \mathcal{L}_{\text{rank}}=\frac{1}{N}\sum_{i<j}\max\big(0,\, m-(s_j-s_i)\big),\quad d_i<d_j,
\end{equation}
and representation geometry is regularized by an InfoNCE-style contrastive loss
\begin{equation}
    \mathcal{L}_{\text{contrast}}=
    -\log \frac{\exp(\langle z_i^{+}, z_i\rangle/\tau)}
    {\sum_k \exp(\langle z_k, z_i\rangle/\tau)} .
\end{equation}
The conditioning token for the denoiser is formed by concatenating the timestep and dose embeddings,
\begin{equation}
    c_t=[e_t;e_d],\qquad e_t=\mathrm{TimeEmbed}(t),
\end{equation}
enabling continuous conditioning under unseen dose levels.

\textbf{2) Prior Extraction (PEM++):} PEM++ enhances structural fidelity by fusing complementary handcrafted and learned priors (Fig.~\ref{fig:PEM}). From $x_t$, multi-scale priors are computed as
\begin{align}
    P_{\text{grad}}&=\nabla x_t,\quad
    P_{\text{lap}}=\Delta x_t, \nonumber\\
    P_{\text{nlm}}&=\mathrm{NLM}(x_t),\quad
    P_{\text{cnn}}=h_{\psi}(x_t),
\end{align}
capturing edges, curvature, non-local similarity, and semantic context. These are combined by attention-based fusion
\begin{equation}
    \alpha_k=\frac{\exp(Q^{T}K_k)}{\sum_j \exp(Q^{T}K_j)},\qquad
    P_{\text{fuse}}=\sum_k \alpha_k V_k,
\end{equation}
with $Q=f_q(x_t)$ and $(K_k,V_k)=f_{k,v}(P_k)$ implemented by $1\times1$ convolutions. The fused prior is modulated by dose and timestep through FiLM:
\begin{equation}
    P'_{\text{fuse}}=\gamma(d,t)\odot P_{\text{fuse}}+\beta(d,t),
\end{equation}
\begin{equation}
    [\gamma(d,t),\beta(d,t)] = W_{\text{film}}[e_d;e_t]+b_{\text{film}},
\end{equation}
and injected into the denoiser,
\begin{equation}
    x'_{t}=U_{\theta}(x_t \oplus P'_{\text{fuse}}, c_t),
\end{equation}
providing dose- and severity-adaptive structural guidance.

\textbf{3) Dose-Calibrated Step Allocation (DCSA):} DCSA adaptively sets the number of reverse refinement steps from the learned dose embedding. A bounded severity indicator is predicted as
\begin{equation}
    \xi_d=\sigma(W_s e_d+b_s),
\end{equation}
and the iteration count is
\begin{equation}
    T(d)=T_{\max}-\left\lfloor \xi_d\,(T_{\max}-T_{\min}) \right\rfloor .
\end{equation}
Thus, lower-dose (more degraded) inputs receive more refinement steps, while near full-dose inputs require fewer, improving efficiency without sacrificing quality.

\subsection{Physics Consistency Enforcement}
To maintain consistency with the measured projections and suppress hallucinated structures, DACD applies a physics-based correction after each diffusion denoising step, enforcing sinogram-domain data fidelity throughout the reverse process.
\paragraph*{Forward-Backprojection Correction}
Let $A$ and $A^{T}$ denote the projection and backprojection operators. Given the intermediate denoised estimate $\tilde{x}_{t-1}$, the projection residual is backprojected to the image domain,
\begin{equation}
    \Delta x_{t-1}=A^{T}\!\big(y_d-A(\tilde{x}_{t-1})\big),
\end{equation}
and used to update the estimate as
\begin{equation}
    x_{t-1}=\tilde{x}_{t-1}+\eta_t\,\Delta x_{t-1},
\end{equation}
where $\eta_t$ controls the correction strength. This step is analogous to one iteration of algebraic reconstruction (SIRT), but is embedded inside every diffusion step, coupling learned denoising with measurement consistency.
\paragraph*{Variational View}
The fixed point of the iterative updates approximately minimizes
\begin{equation}
    \hat{x}=\arg\min_x
    \|A(x)-y_d\|_2^2
    +\lambda_{\text{p}}\|x-U_{\theta}(x,c)\|_2^2,
\end{equation}
which balances projection-domain data fidelity with adherence to the diffusion prior $U_{\theta}$. The first term enforces agreement with the measured sinogram, while the second keeps the solution close to the learned denoising manifold.

\begin{table*}[!ht]
\centering
\footnotesize
\setlength{\tabcolsep}{3pt}
\renewcommand{\arraystretch}{1.05}
\caption{Quantitative comparison on Mayo-2020 across anatomies and dose levels.
Best results are in \textbf{bold} and second-best are \underline{underlined}.
$^\ast$ indicates a significant difference from the best method ($p<0.05$, Wilcoxon test).}
\label{tab:mayo2020_sota_per_anatomy_dose}

\resizebox{\textwidth}{!}{%
\begin{tabular}{llcccc}
\toprule
\textbf{Anatomy} & \textbf{Method} &
\multicolumn{1}{c}{\textbf{50\%}} &
\multicolumn{1}{c}{\textbf{25\%}} &
\multicolumn{1}{c}{\textbf{12.5\%}} &
\multicolumn{1}{c}{\textbf{5\%}} \\
\cmidrule(lr){3-6}
& & \multicolumn{1}{c}{PSNR / SSIM / RMSE}
  & \multicolumn{1}{c}{PSNR / SSIM / RMSE}
  & \multicolumn{1}{c}{PSNR / SSIM / RMSE}
  & \multicolumn{1}{c}{PSNR / SSIM / RMSE} \\
\midrule

\multirow{8}{*}{\textbf{Abdomen}}
& PrideDiff$^\ast$      & \underline{41.24} / \underline{0.9748} / \underline{0.013} & \underline{39.83} / \underline{0.970} / \underline{0.016} & \underline{37.92} / \underline{0.963} / \underline{0.020} & \underline{35.45} / \underline{0.955} / \underline{0.025} \\
& Cold Diffusion$^\ast$ & 39.36 / 0.969 / 0.015 & 38.07 / 0.964 / 0.018 & 36.18 / 0.957 / 0.022 & 34.01 / 0.948 / 0.028 \\
& CoreDiff$^\ast$       & 40.31 / 0.972 / 0.014 & 38.94 / 0.967 / 0.017 & 37.17 / 0.961 / 0.021 & 34.83 / 0.952 / 0.026 \\
& RDDM$^\ast$           & 39.75 / 0.970 / 0.015 & 38.32 / 0.965 / 0.018 & 36.63 / 0.958 / 0.022 & 34.34 / 0.949 / 0.027 \\
& DDPM-1000$^\ast$      & 39.05 / 0.967 / 0.016 & 37.64 / 0.962 / 0.019 & 35.82 / 0.955 / 0.023 & 33.63 / 0.946 / 0.029 \\
& RED-diff$^\ast$       & 38.57 / 0.964 / 0.017 & 37.14 / 0.959 / 0.020 & 35.36 / 0.952 / 0.024 & 33.23 / 0.943 / 0.030 \\
& Noise2Sim$^\ast$      & 38.08 / 0.962 / 0.018 & 36.62 / 0.956 / 0.021 & 34.93 / 0.949 / 0.025 & 32.85 / 0.940 / 0.031 \\
& \textbf{DACD (Ours)}  & \textbf{42.02} / \textbf{0.9773} / \textbf{0.012} & \textbf{40.57} / \textbf{0.982} / \textbf{0.015} & \textbf{38.64} / \textbf{0.966} / \textbf{0.019} & \textbf{36.23} / \textbf{0.958} / \textbf{0.024} \\
\midrule

\multirow{8}{*}{\textbf{Chest}}
& PrideDiff$^\ast$      & \underline{38.33} / \underline{0.8719} / \underline{0.018} & \underline{36.95} / \underline{0.862} / \underline{0.021} & \underline{35.03} / \underline{0.853} / \underline{0.025} & \underline{32.72} / \underline{0.842} / \underline{0.030} \\
& Cold Diffusion$^\ast$ & 36.83 / 0.866 / 0.020 & 35.41 / 0.857 / 0.023 & 33.52 / 0.847 / 0.027 & 31.42 / 0.836 / 0.032 \\
& CoreDiff$^\ast$       & 37.63 / 0.869 / 0.019 & 36.20 / 0.860 / 0.022 & 34.34 / 0.851 / 0.026 & 32.03 / 0.840 / 0.031 \\
& RDDM$^\ast$           & 37.12 / 0.868 / 0.019 & 35.74 / 0.858 / 0.022 & 33.93 / 0.849 / 0.026 & 31.72 / 0.838 / 0.031 \\
& DDPM-1000$^\ast$      & 36.46 / 0.864 / 0.020 & 35.03 / 0.855 / 0.023 & 33.14 / 0.846 / 0.027 & 30.94 / 0.835 / 0.032 \\
& RED-diff$^\ast$       & 35.87 / 0.861 / 0.021 & 34.45 / 0.852 / 0.024 & 32.62 / 0.843 / 0.028 & 30.43 / 0.832 / 0.033 \\
& Noise2Sim$^\ast$      & 35.31 / 0.858 / 0.022 & 33.91 / 0.849 / 0.025 & 32.11 / 0.840 / 0.029 & 29.84 / 0.829 / 0.034 \\
& \textbf{DACD (Ours)}  & \textbf{39.02} / \textbf{0.8768} / \textbf{0.017} & \textbf{37.61} / \textbf{0.869} / \textbf{0.020} & \textbf{35.84} / \textbf{0.861} / \textbf{0.024} & \textbf{33.53} / \textbf{0.850} / \textbf{0.029} \\
\midrule

\multirow{8}{*}{\textbf{Head}}
& PrideDiff$^\ast$      & \underline{44.83} / \underline{0.9794} / \underline{0.011} & \underline{43.11} / \underline{0.975} / \underline{0.013} & \underline{41.03} / \underline{0.970} / \underline{0.016} & \underline{38.23} / \underline{0.963} / \underline{0.021} \\
& Cold Diffusion$^\ast$ & 43.04 / 0.974 / 0.012 & 41.44 / 0.969 / 0.015 & 39.32 / 0.964 / 0.018 & 36.63 / 0.957 / 0.023 \\
& CoreDiff$^\ast$       & 44.05 / 0.977 / 0.012 & 42.35 / 0.972 / 0.014 & 40.21 / 0.967 / 0.017 & 37.52 / 0.960 / 0.022 \\
& RDDM$^\ast$           & 43.42 / 0.976 / 0.012 & 41.83 / 0.970 / 0.015 & 39.84 / 0.965 / 0.018 & 37.03 / 0.958 / 0.023 \\
& DDPM-1000$^\ast$      & 42.61 / 0.972 / 0.013 & 40.94 / 0.967 / 0.016 & 38.86 / 0.961 / 0.019 & 36.15 / 0.954 / 0.024 \\
& RED-diff$^\ast$       & 42.04 / 0.969 / 0.014 & 40.34 / 0.964 / 0.017 & 38.32 / 0.958 / 0.020 & 35.54 / 0.951 / 0.025 \\
& Noise2Sim$^\ast$      & 41.52 / 0.967 / 0.015 & 39.83 / 0.962 / 0.018 & 37.84 / 0.956 / 0.021 & 35.03 / 0.949 / 0.026 \\
& \textbf{DACD (Ours)}  & \textbf{46.03} / \textbf{0.9821} / \textbf{0.010} & \textbf{44.35} / \textbf{0.978} / \textbf{0.012} & \textbf{42.24} / \textbf{0.972} / \textbf{0.015} & \textbf{39.42} / \textbf{0.965} / \textbf{0.020} \\
\bottomrule
\end{tabular}}
\end{table*}
\section{Experimental Setup and Implementation}
\subsection{Datasets}
We evaluate on Mayo-2016~\cite{mccollough2017low}, Mayo-2020~\cite{moen2021low}, and LoDoPaB-CT~\cite{leuschner2021lodopab}. 
Mayo-2020 contains 300 full-dose volumetric scans (head, chest, abdomen). All DICOMs were converted to Hounsfield Units (HU) and clipped to $[-1024,3072]$ HU. Clean parallel-beam projections were generated from full-dose images, and low-dose data were simulated by Poisson thinning at $d\in\{50\%,25\%,12.5\%,5\%\}$, followed by FBP to obtain inputs $x_{\mathrm{fbp}}(d)$. For Mayo-2016, provided quarter-dose scans were used as low-dose inputs and full-dose scans as ground truth with the same HU preprocessing. 
LoDoPaB-CT is a large-scale thoracic benchmark (33{,}525/3{,}552/3{,}528 train/val/test slices) with simulated low-dose parallel-beam projections; inputs were obtained via FBP from the provided sinograms and HU-clipped to the same range. Its scale and pulmonary variability provide a challenging test of generalization.

\subsection{Implementation Details}
All models were implemented in \texttt{PyTorch} with CUDA acceleration. Forward and backprojection were performed using the GPU-accelerated, differentiable \texttt{TorchRadon} library. Unless stated otherwise, filtered backprojection (FBP) used a Ram-Lak filter. CT slices were cropped to $384\times384$ patches to balance memory usage and context. Training used the AdamW optimizer with learning rate $1\times10^{-4}$, $\beta_1=0.9$, $\beta_2=0.999$, and weight decay $10^{-2}$, together with cosine annealing and linear warm-up. The batch size was $8$ per GPU. Models were trained for $100$ epochs in Stage~A and $200$ epochs in Stage~B on four NVIDIA RTX~4090 GPUs with automatic mixed precision enabled.

\subsection{Evaluation Metrics}
Performance was measured using PSNR, SSIM, and RMSE between reconstruction $\hat{x}$ and ground truth $x_{\mathrm{gt}}$:
\begin{align}
\mathrm{PSNR} &= 10\log_{10}\!\left(\frac{L^2}{\|\hat{x}-x_{\mathrm{gt}}\|_2^2/N}\right),\\
\mathrm{SSIM} &= 
\frac{(2\mu_{\hat{x}}\mu_{x_{\mathrm{gt}}}+C_1)(2\sigma_{\hat{x}x_{\mathrm{gt}}}+C_2)}
{(\mu_{\hat{x}}^2+\mu_{x_{\mathrm{gt}}}^2+C_1)(\sigma_{\hat{x}}^2+\sigma_{x_{\mathrm{gt}}}^2+C_2)},\\
\mathrm{RMSE} &= \sqrt{\frac{1}{N}\|\hat{x}-x_{\mathrm{gt}}\|_2^2},
\end{align}
where $L$ is the HU dynamic range and $N$ the number of pixels. Higher PSNR/SSIM and lower RMSE indicate better quality. Metrics were computed per slice. Statistical significance was assessed using paired Wilcoxon signed-rank tests with FDR control $(q=0.05)$.
\begin{figure*}[!ht]
\centering
\includegraphics[width=1\linewidth]{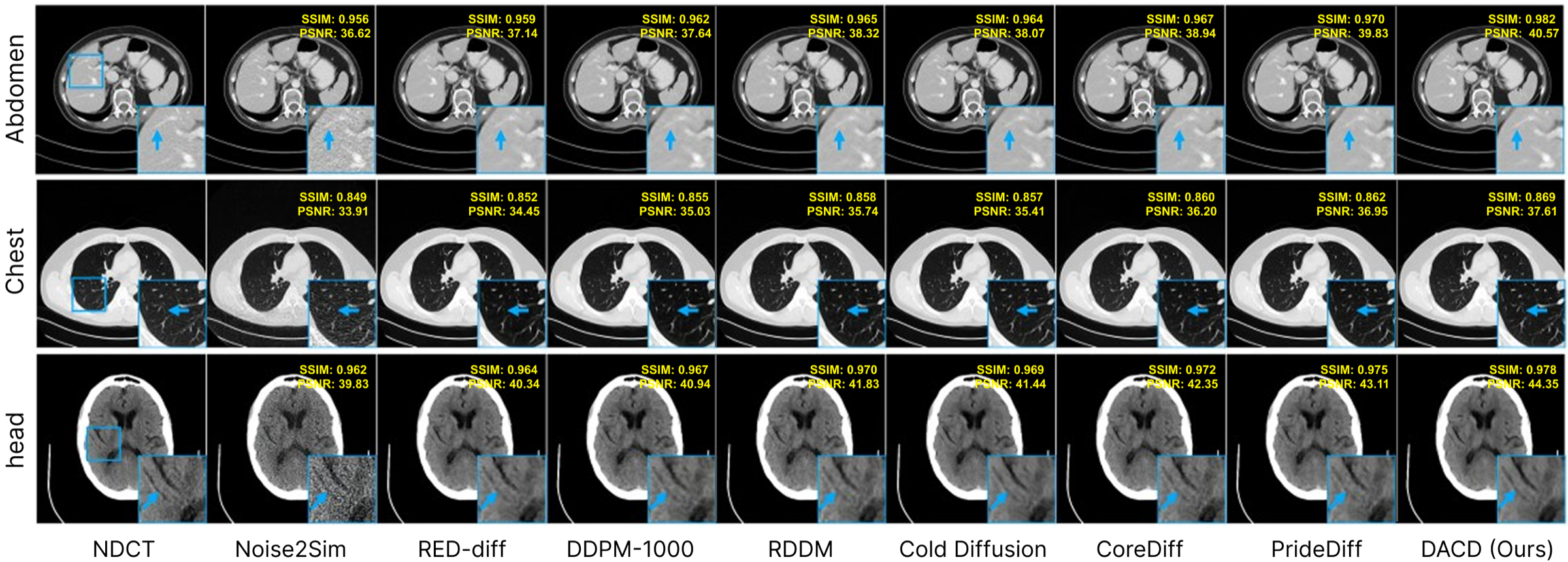}
\caption{Qualitative comparison of reconstructed CT images from different methods on the Mayo 2020 dataset. Zoomed-in regions highlight that the proposed DACD framework achieves better texture and noise suppression compared to other approaches.}
\label{fig:mayo2020_Qualitative}
\end{figure*}
\begin{figure*}[!ht]
    \centering
    \includegraphics[width=1\linewidth]{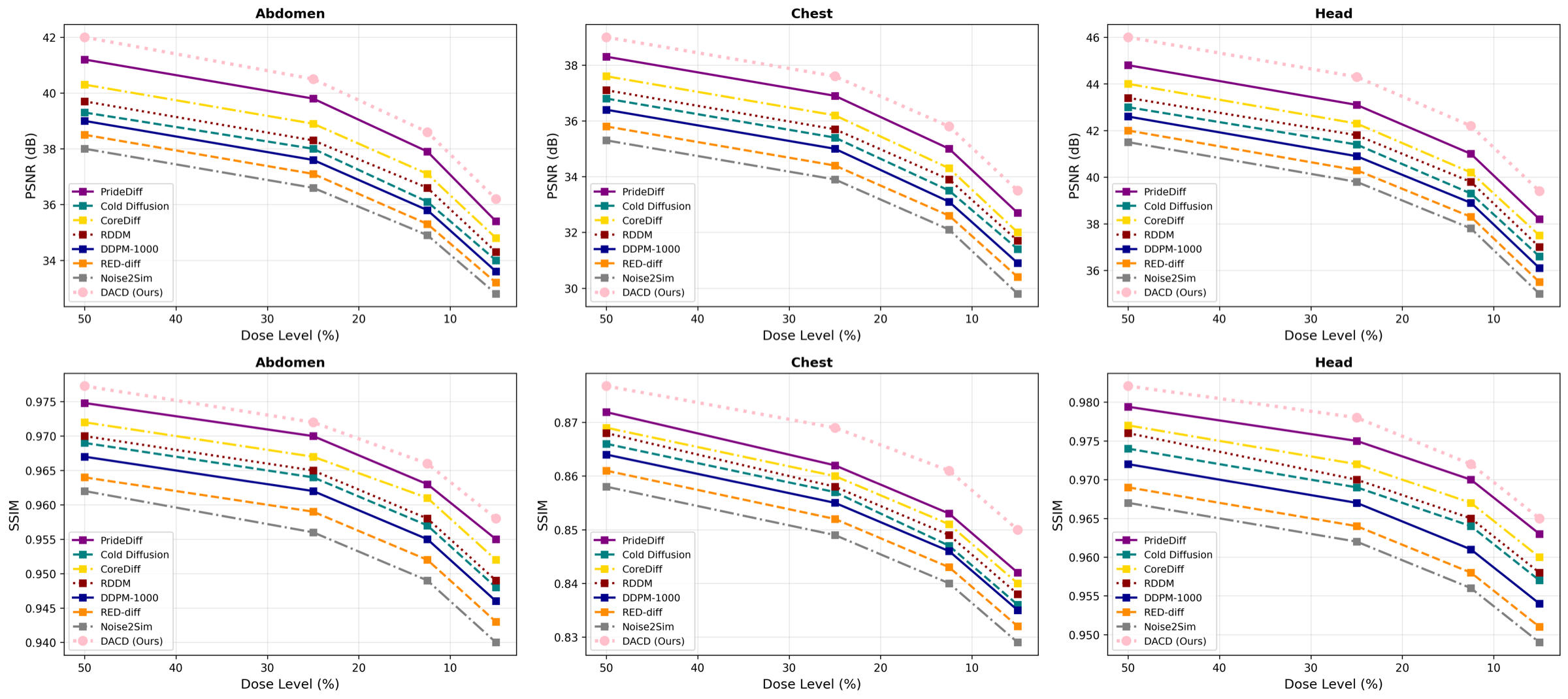}
    \caption{Quantitative performance across anatomical regions and dose levels on Mayo-2020. The proposed DACD (pink line) consistently achieves the highest PSNR and SSIM across anatomical regions, reflecting robust dose adaptability and structural fidelity under severe noise conditions.}
    \label{fig:mayo20Linechart}
\end{figure*}
\section{Results and Analysis}
\subsection{Results on the Mayo-2020 Dataset}
DACD achieves the best performance across all anatomies and dose levels. In the abdomen, it improves PSNR by +0.78\,dB over the second-best method at both 50\% and 5\% dose (42.02 vs.\ 41.24\,dB and 36.23 vs.\ 35.45\,dB), corresponding to up to 7.7\% lower RMSE. In the chest, gains reach +0.81\,dB at 5\% dose (33.53 vs.\ 32.72\,dB) with approximately 5-6\% RMSE reduction. The largest improvements are observed in the head region, where DACD provides up to +1.20\,dB PSNR at 50\% dose and +1.19\,dB at 5\% dose, together with up to 9.1\% lower RMSE. Consistent advantages are also observed at intermediate doses, with SSIM improvements of up to +0.008. These results demonstrate stable and statistically significant gains across anatomies and noise levels, confirming the robustness of the proposed dose-aware and physics-consistent reconstruction framework. And figure.~\ref{fig:mayo2020_Qualitative} provides a visual comparison across three representative regions abdomen, chest, and head. DACD preserves subtle texture and vessel details with minimal oversmoothing, whereas competing approaches (DDPM-1000~\cite{choi2021ilvr}, RDDM~\cite{liu2024residual}, Cold Diffusion~\cite{yen2023cold}) exhibit visible blurring or loss of low-contrast boundaries.

\subsection{Cross-Dataset Evaluation on LoDoPaB-CT and Mayo-2016}
Table~\ref{tab:combined_results} reports quantitative results on two out-of-distribution benchmarks, LoDoPaB-CT and Mayo-2016. On LoDoPaB-CT, DACD achieves the best accuracy with 38.74\,dB PSNR and 0.8951 SSIM, surpassing PrideDiff (37.92\,dB / 0.891) and CoreDiff (37.40\,dB / 0.888), while obtaining the lowest RMSE of 0.0281. 
This corresponds to PSNR gains of +0.82\,dB ($\approx$2.2\%) over PrideDiff and +1.34\,dB ($\approx$3.6\%) over CoreDiff, together with RMSE reductions of about 4.1\% and 5.7\%, demonstrating robustness to unseen dose distributions and acquisition geometries. A similar pattern is observed on Mayo-2016, where DACD reaches 44.36\,dB PSNR and 0.9724 SSIM, outperforming PrideDiff (43.72\,dB / 0.9693) and CoreDiff (43.25\,dB / 0.9670). 
These correspond to PSNR improvements of +0.64\,dB ($\approx$1.5\%) and +1.11\,dB ($\approx$2.6\%), along with the lowest RMSE of 0.0243, indicating higher fidelity and lower variance across slices. 
The reduced dispersion suggests more stable and reliable reconstructions under diverse anatomical and noise conditions, consistent with clinical-quality imaging requirements.
\begin{figure*}[!ht]
\centering
\includegraphics[width=1\linewidth]{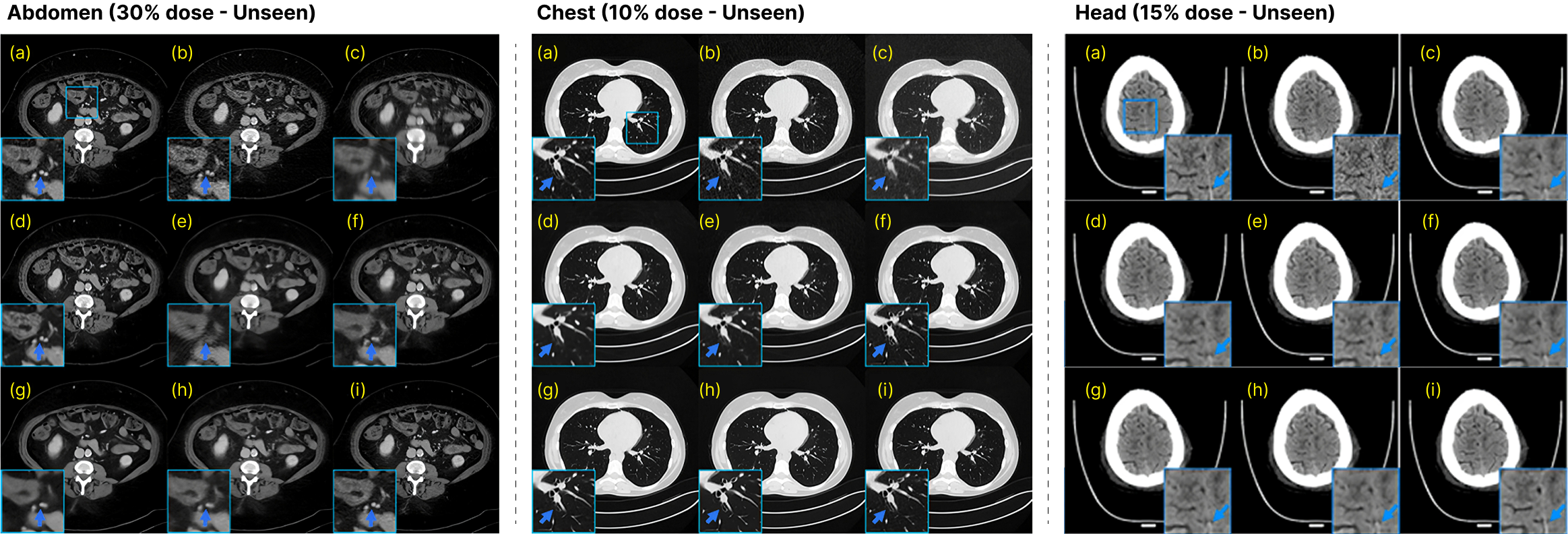}
\caption{Visual results on unseen dose levels across abdomen (30\%), chest (10\%), and head (15\%) cases. DACD (h) maintains sharper tissue boundaries and more realistic contrast than competing methods, demonstrating strong generalization to dose conditions not encountered during training.}
\label{fig:unseen_dose}
\end{figure*}

\begin{table*}[!ht]
\centering
\caption{Quantitative results on LoDoPaB-CT and Mayo-2016. Mean~$\pm$~standard deviation of PSNR~(dB), SSIM, and RMSE. Bold and underlined values denote the best and second-best results. * indicates difference from the best method ($p < 0.05$, Wilcoxon test).}
\label{tab:combined_results}
\resizebox{\textwidth}{!}{
\begin{tabular}{lcccccc}
\toprule
\multirow{2}{*}{\textbf{Method}} & \multicolumn{3}{c}{\textbf{LoDoPaB-CT}} & \multicolumn{3}{c}{\textbf{Mayo-2016}} \\
\cmidrule(lr){2-4} \cmidrule(lr){5-7}
 & \textbf{PSNR (dB)} & \textbf{SSIM} & \textbf{RMSE} & \textbf{PSNR (dB)} & \textbf{SSIM} & \textbf{RMSE} \\
\midrule
Cold Diffusion   & 36.10 $\pm$ 0.41* & 0.883 $\pm$ 0.0041* & 0.0313 $\pm$ 0.0009* & 41.85 $\pm$ 0.42* & 0.9632 $\pm$ 0.0031* & 0.0278 $\pm$ 0.0009* \\
RDDM             & 36.25 $\pm$ 0.38* & 0.885 $\pm$ 0.0038* & 0.0310 $\pm$ 0.0008* & 42.10 $\pm$ 0.38* & 0.9641 $\pm$ 0.0028* & 0.0273 $\pm$ 0.0008* \\
DDPM-1000        & 35.85 $\pm$ 0.45* & 0.880 $\pm$ 0.0043* & 0.0318 $\pm$ 0.0010* & 41.50 $\pm$ 0.47* & 0.9610 $\pm$ 0.0035* & 0.0281 $\pm$ 0.0010* \\
RED-diff         & 35.55 $\pm$ 0.48* & 0.876 $\pm$ 0.0046* & 0.0324 $\pm$ 0.0011* & 40.95 $\pm$ 0.49* & 0.9582 $\pm$ 0.0039* & 0.0290 $\pm$ 0.0011* \\
Noise2Sim        & 35.05 $\pm$ 0.50* & 0.871 $\pm$ 0.0048* & 0.0333 $\pm$ 0.0012* & 40.40 $\pm$ 0.52* & 0.9547 $\pm$ 0.0041* & 0.0301 $\pm$ 0.0013* \\
CoreDiff         & 37.40 $\pm$ 0.34* & 0.888 $\pm$ 0.0035* & 0.0298 $\pm$ 0.0007* & 43.25 $\pm$ 0.33* & 0.9670 $\pm$ 0.0025* & 0.0256 $\pm$ 0.0007* \\
\underline{PrideDiff} & \underline{37.92 $\pm$ 0.29*} & \underline{0.891 $\pm$ 0.0031*} & \underline{0.0293 $\pm$ 0.0006*} & \underline{43.72 $\pm$ 0.29*} & \underline{0.9693 $\pm$ 0.0023*} & \underline{0.0251 $\pm$ 0.0006*} \\
\textbf{DACD (Ours)} & \textbf{38.74 $\pm$ 0.26} & \textbf{0.8951 $\pm$ 0.0029} & \textbf{0.0281 $\pm$ 0.0005} & \textbf{44.36 $\pm$ 0.27} & \textbf{0.9724 $\pm$ 0.0021} & \textbf{0.0243 $\pm$ 0.0005} \\
\bottomrule
\end{tabular}}
\end{table*}
\subsection{Results on Unseen Dose Levels}
To assess extrapolation to dose levels not seen during training, we evaluate at 30\% (abdomen), 10\% (chest), and 15\% (head). DACD consistently achieves the highest PSNR/SSIM and lowest RMSE across all cases. 
At the unseen 10\% chest dose, DACD attains 37.8\,dB PSNR and 0.870 SSIM, clearly surpassing PrideDiff and CoreDiff, indicating strong robustness to dose mismatch. Across regions, DACD preserves fine structures while suppressing noise-induced streaks, as illustrated in Fig.~\ref{fig:unseen_dose}. These results show that dose-aware conditioning with in-loop physics consistency enables reliable generalization beyond the trained dose range. While diffusion baselines such as CoreDiff and RDDM remain competitive at moderate noise, conventional denoisers (e.g., RED-CNN~\cite{chen2017low} and Noise2Sim~\cite{niu2022noise}) exhibit larger drops, particularly in SSIM, highlighting limited cross-dose generalization.

\subsection{Ablation Study}
Table~\ref{tab:ablation_components} quantifies the contribution of each component on Mayo-2020 at 12.5\% dose. 
Starting from the cold diffusion baseline (37.10\,dB / 0.961), adding the dose-aware perception (DAP) module yields a +0.80\,dB PSNR gain ($\approx$2.2\%) and a +0.004 SSIM increase. 
Incorporating the structural prior module (PEM++) further improves PSNR to 38.70\,dB, providing an additional +0.80\,dB ($\approx$2.1\%) and raising SSIM to 0.968. 
Introducing dose-calibrated step allocation (DCSA) boosts performance to 39.25\,dB and 0.970, corresponding to a cumulative +2.15\,dB ($\approx$5.8\%) improvement over the baseline. 
Finally, adding the physics-consistency (PC) correction achieves the best result of 39.80\,dB / 0.972, giving a total gain of +2.70\,dB ($\approx$7.3\%) and +0.011 SSIM relative to the baseline. 
These monotonic improvements demonstrate that each module contributes complementary benefits, with the largest cumulative effect arising from the combination of dose awareness, structural priors, adaptive step allocation, and projection-domain consistency.
\begin{table}[!ht]
\centering
\small
\setlength{\tabcolsep}{6pt}
\renewcommand{\arraystretch}{1.08}
\caption{Ablation study on the \textbf{Mayo-2020} dataset at 12.5\% dose.}
\label{tab:ablation_components}

\begin{tabular}{>{\raggedright\arraybackslash}p{0.58\columnwidth}cc}
\toprule
\textbf{Configuration} & \textbf{PSNR (dB)} & \textbf{SSIM} \\
\midrule
Cold Diffusion (baseline) & 37.10 & 0.961 \\
+ DAP & 37.90 & 0.965 \\
+ DAP + PEM++ & 38.70 & 0.968 \\
+ DAP + PEM++ + DCSA & 39.25 & 0.970 \\
\textbf{Full DACD (DAP + PEM++ + DCSA + PC)} & \textbf{39.80} & \textbf{0.972} \\
\bottomrule
\end{tabular}
\end{table}
\section{Conclusion}
This work presented DACD, a physics-consistent framework for generalizable low-dose CT reconstruction. By modeling radiation dose as a continuous latent factor and integrating dose-aware perception, dose-calibrated step allocation, and in-loop forward-backprojection correction, DACD achieves stable and reliable reconstruction across a wide range of dose levels, including previously unseen conditions. Experimental results on multiple public benchmarks demonstrate that the proposed framework consistently improves reconstruction accuracy and structural fidelity under severe noise. These findings indicate that jointly enforcing dose-aware conditioning and projection-domain consistency within the diffusion process is an effective strategy for robust low-dose CT reconstruction. Future work will explore anatomy-aware representations and further improve cross-scanner generalization and inference efficiency to facilitate practical clinical deployment.

\bibliographystyle{IEEEtran}
\bibliography{references}

\end{document}